\documentclass[letterpaper, 10 pt, conference]{ieeeconf}  

\IEEEoverridecommandlockouts                              

\usepackage[noadjust]{cite}
\usepackage{color}
\usepackage{lipsum}
\usepackage{siunitx}
\usepackage{multirow} 
\usepackage{graphicx}
\usepackage{caption}
\usepackage{tabularx}
\usepackage{siunitx}
\usepackage{threeparttable}
\usepackage{threeparttablex}
\usepackage{bbm}
\usepackage{float}

\let\labelindent\relax

\usepackage{enumitem} 

\usepackage{amsmath}
\usepackage{amssymb}
\usepackage[
linesnumbered,ruled,vlined]{algorithm2e}
\usepackage{algpseudocode}
\usepackage{algorithmicx}
\usepackage{algcompatible}

\title{\LARGE \bf
Efficient Navigation Among Movable Obstacles \\ using a Mobile Manipulator via Hierarchical Policy Learning
}

\author{Taegeun Yang\textsuperscript{1}, Jiwoo Hwang\textsuperscript{2}, Jeil Jeong\textsuperscript{2}, Minsung Yoon\textsuperscript{1} and Sung-Eui Yoon\textsuperscript{1}\textsuperscript{\textdagger}%
\thanks{\textsuperscript{1}T. Yang, M. Yoon, and S. Yoon are with the School of Computing in Korea Advanced Institute of Science and Technology (KAIST),
        Daejeon, 34141, Republic of Korea.}%
\thanks{\textsuperscript{2}J. Jeong and J. Hwang are with the Robotics Program at the same institute, KAIST.}%
\thanks{\textsuperscript{\textdagger}S. Yoon is a corresponding author; {\tt\small sungeui@kaist.edu}.}
}

\begin{document}

\maketitle
\thispagestyle{empty}
\pagestyle{empty}

    \begin{abstract}
We propose a hierarchical reinforcement learning (HRL) framework for efficient Navigation Among Movable Obstacles (\textit{NAMO}) using a mobile manipulator. 
Our approach combines interaction-based obstacle property estimation with structured pushing strategies, facilitating the dynamic manipulation of unforeseen obstacles while adhering to a pre-planned global path. 
The high-level policy generates pushing commands that consider environmental constraints and path-tracking objectives, while the low-level policy precisely and stably executes these commands through coordinated whole-body movements.
Comprehensive simulation-based experiments demonstrate improvements in performing \textit{NAMO} tasks, including higher success rates, shortened traversed path length, and reduced goal-reaching times, compared to baselines.
Additionally, ablation studies assess the efficacy of each component, while a qualitative analysis further validates the accuracy and reliability of the real-time obstacle property estimation.

\end{abstract}

    \section{Introduction}
\begin{figure}[t]
    \vspace{0.2cm}
  \centering
  \includegraphics[width=0.49\textwidth]{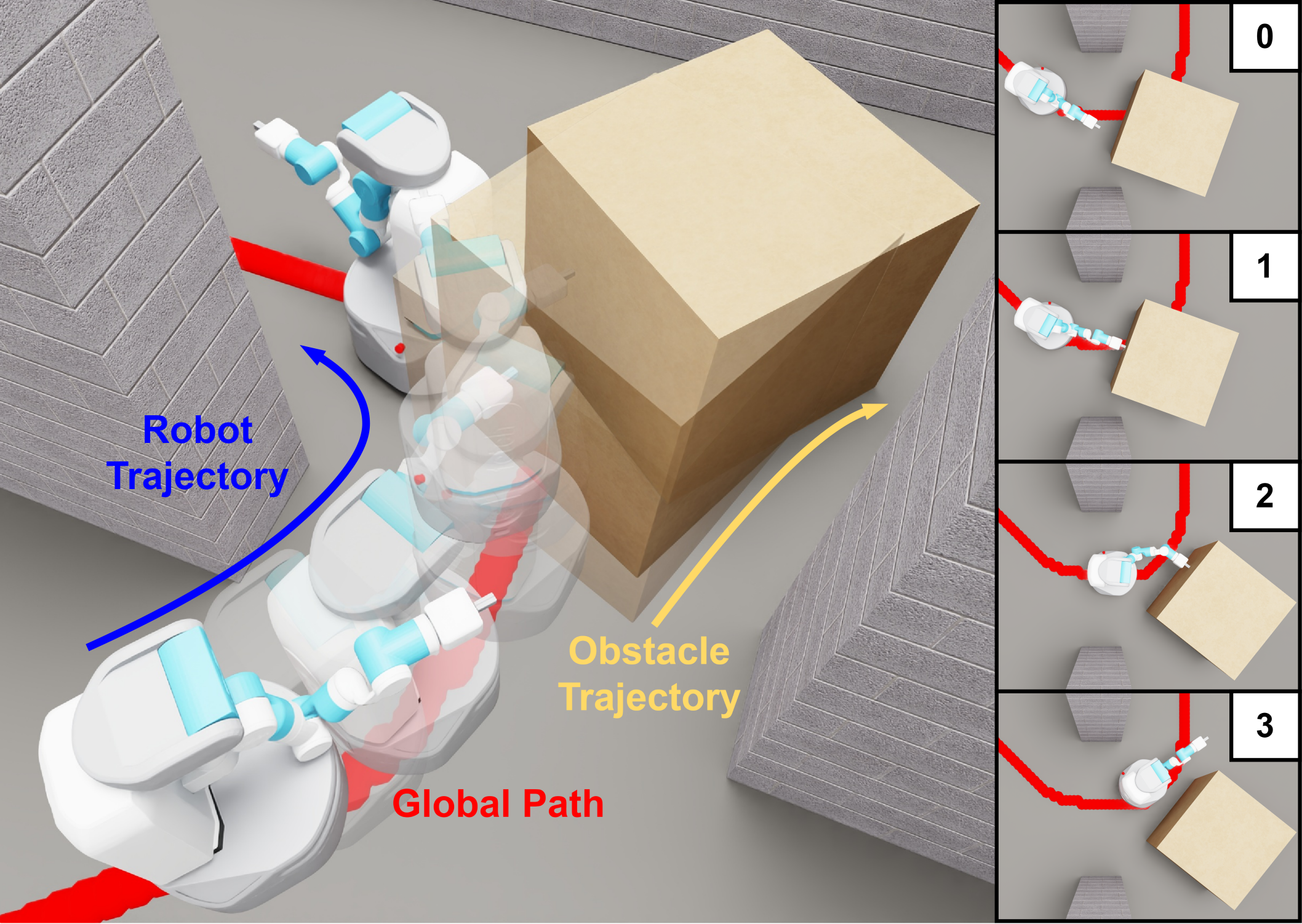}
  \caption{\small \textbf{Whole-Body Mobile Manipulator Control for \textit{NAMO}}. Illustration of the \textit{NAMO} task, where the mobile manipulator follows a global path (red) while actively clearing a blocking obstacle using whole-body coordinated motions.
  During the interaction, the robot continuously tracks the global path with its base, while simultaneously pushing the obstacle aside to clear the way using its manipulator arm.
  The right sequence shows the interaction process between the robot and the obstacle.
  }
  \label{fig:fig_1}
  \vspace{-0.6cm}
\end{figure}

Robust robot navigation in complex environments is crucial for applications ranging from delivery~\cite{kim2024development} to warehouse automation~\cite{lee2021mobile}.
While recent methods excel at collision avoidance, they could fail when physical object manipulation is necessary to create feasible paths, for instance, when narrow passages are blocked by obstacles, as shown in Fig.~\ref{fig:fig_1}.

Navigation Among Movable Obstacles (\textit{NAMO}) research addresses this challenge by enabling robots to manipulate objects (i.e., movable obstacles) to actively create navigable regions. 
Conventional offline \textit{NAMO} methods require complete environmental knowledge, whereas recent online approaches operate with minimal or without global information, such as floor plans that only include large structures like walls, to generate coarse global paths and dynamically respond to encountered obstacles during navigation. (refer to Sec.~\ref{sec:related-A})

By leveraging the advancements in reinforcement learning (RL), some researchers in \textit{NAMO} tasks have enhanced decision-making processes for robots, enabling efficient navigation while manipulating objects when necessary.
However, existing research overlooks physical attributes of movable obstacles, such as mass, friction, and center of mass, which are critical for effective manipulation. 
To address this limitation, we adopt real-time property estimation approaches, allowing robots to better understand and interact with objects. 
This integration enhances manipulation accuracy and eventually improves navigation efficiency. (refer to Sec.~\ref{sec:related-B})

Furthermore, we employ hierarchical reinforcement learning (HRL) frameworks to manage the complexity of \textit{NAMO} tasks. 
By decoupling high-level decision-making from low-level motor execution, HRL allows robots to generate strategic commands, such as determining where to push a blocking obstacle with the manipulator's end-effector, and how quickly to move the base at a specific angle, as illustrated in Fig.~\ref{fig:fig_1}. 
These commands are then executed with precision through a low-level controller.
This hierarchical structure not only streamlines the decision-making process but also enables more efficient and adaptive behaviors in complex, unstructured environments, resulting in improved overall \textit{NAMO} task performance. (refer to Sec.~\ref{sec:related-C})

\noindent In this work, our main contributions are \textit{threefold}:
\begin{itemize}[leftmargin=0.4cm]
    \item A HRL framework that integrates mobile navigation with obstacle manipulation using a seven degrees-of-freedom (DOFs) mobile manipulator, the Fetch, to enable efficient navigation in unstructured environments for \textit{NAMO} tasks;
    \item An interaction-based property estimation module that facilitates adaptive interaction with diverse movable objects;
    \item Comprehensive simulation-based experiments demonstrating improved navigation performance, in terms of success rates, path length, and goal-reaching times, along with ablation studies assessing the effectiveness of each component and qualitative analysis of the property estimation.
\end{itemize}

    \section{Related Work}

\subsection{Navigation Among Movable Obstacles (NAMO)}
\label{sec:related-A}
\textit{NAMO} represents robot navigation tasks in environments where obstacles need to be physically manipulated to create traversable paths toward goal locations. 
Traditional approaches~\cite{wilfong1988motion, stilman2008planning, stilman2005navigation, okada2004environment} assume complete environmental knowledge, including floor plans and detailed obstacle information (e.g., poses, geometries, movability), limiting their applicability in dynamic and unexplored environments.
To mitigate this requirement, recent online methods~\cite{wang2020affordance, wu2010navigation, levihn2014locally, levihn2013hierarchical, raghavan2021reconfigurable, ellis2022navigation, ellis2023navigation} utilize approximate global paths from coarse floor maps or even operate with only local sensing information, dynamically responding to unexpected obstacles during navigation toward goals.

Recent advancements in reinforcement learning (RL) have enabled optimal decision-making in complex \textit{NAMO} tasks that require simultaneous navigation and object manipulation~\cite{zeng2021pushing, wang2023curriculum}.
However, existing RL-based methods overlook the physical properties of obstacles during interaction, such as mass, friction, or center of mass, often resulting in inaccurate manipulation and reduced navigation efficiency, as shown in Sec.~\ref{sec:exp}. 
In contrast, we incorporate real-time property estimation within our framework, improving obstacle manipulation and ultimately enhancing navigation efficiency.

\subsection{Interaction-Based Estimation of Object Properties} 
\label{sec:related-B}
Accurate estimation of physical properties is essential for various domains, including system simulation~\cite{raissi2020hidden}, legged locomotion~\cite{kim2025online}, and object manipulation~\cite{lee2024accurate}. 
Properties, such as mass, center of mass, and friction, significantly influence motion and interaction dynamics, making their estimation crucial for effective robotic control.
While some objects have well-defined and predictable properties, real-world environments often contain unknown or visually similar objects with differing physical characteristics, requiring robots to infer these properties through direct interaction.

In object manipulation, traditional analytical approaches often struggle with real-world uncertainties and object variability~\cite{yu1999estimation, tanaka2004active, yu2005estimation, methil2006pushing}. 
To address these limitations, recent methods utilize large-scale interaction data and learning-based inference for accurate and robust object property estimation~\cite{yu2017preparing, xu2019densephysnet, mavrakis2020estimating}. 
Building on these advancements, we incorporate property estimation into RL frameworks for \textit{NAMO} tasks, enabling more context-aware manipulation policies.

\begin{figure*}[t]
  \centering
  \includegraphics[width=2\columnwidth]{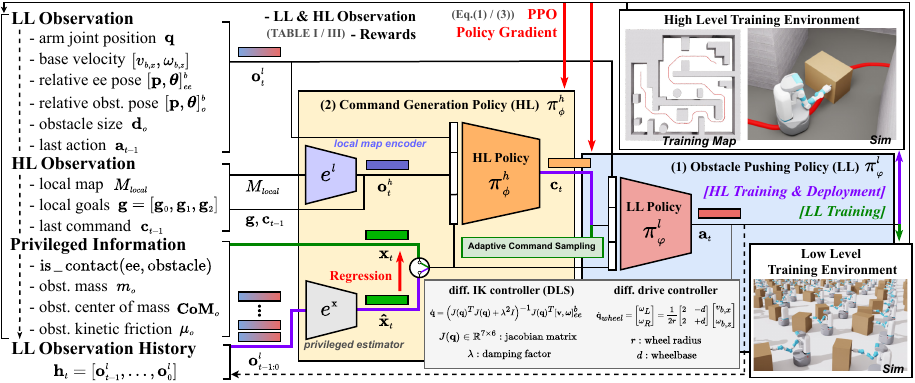}
  \caption{\small \textbf{Training Framework}.
  Our framework consists of two stages:
  (1) The low-level policy learns to execute whole-body actions that achieve the commanded robot-obstacle kinematic configuration (i.e., pushing command) while maintaining stable contact to push obstacles (refer to Sec.~\ref{sec:command}).
  An adaptive command sampling strategy is used to train the policy, enabling it to handle a broader range of commands (Sec.~\ref{sec:curriculum}).
  (2) The high-level policy is trained to generate pushing commands for low-level that guide the robot's base along the planned path while clearing obstacles, considering environmental factors.
  A history encoder is trained to estimate contact state and obstacle properties from low-level observation histories (Sec.~\ref{sec:priv_estimate}).
  The term 'obst.' in the figure denotes 'obstacle' for brevity.
  }
  \label{fig:framework}
  \vspace{-0.6cm}
\end{figure*}

\subsection{Hierarchical Reinforcement Learning}
\label{sec:related-C}
Hierarchical control structures simplify complex robotic tasks by decoupling high-level decision-making from low-level execution~\cite{lee2024learning, miki2024learning, xu2024dexterous, li2020hrl4in, hong2024learning, xia2020relmogen}. 
For instance, in navigation, high-level controllers generate body velocity independent of the robot's specific embodiment~\cite{lee2024learning, miki2024learning, xu2024dexterous}. 
Likewise, in manipulation, high-level controllers determine interaction strategies, such as motor skill selection~\cite{li2020hrl4in} and sub-goal generation~\cite{hong2024learning, xia2020relmogen}, while low-level controllers execute high-level commands to generate precise motions for completing tasks.

To resolve the complexity of \textit{NAMO} tasks, we adopt the hierarchical structure, where the high-level command space is defined as a pushing command space (refer to Sec.~\ref{sec:command}).
The high-level controller generates pushing commands to accurately track the planned global path while clearing an obstacle through manipulation, and the low-level controller translates the high-level pushing commands into coordinated whole-body movements.
Our hierarchical framework enhances navigation efficiency compared to an end-to-end structure, as exhibited in Sec.~\ref{sec:exp}.

    \section{Variable Notation}
This section defines the variable notation used throughout this manuscript.
Vectors \(\mathbf{p}, \mathbf{v} \), and \( \dot{\mathbf{v}} \in \mathbb{R}^3\) represent position, velocity, and acceleration in Cartesian space.
Orientation is expressed using Euler angles in the XYZ convention as \(\boldsymbol{\theta} \in \mathbb{R}^3\), with \(\boldsymbol{\omega}\) and \(\dot{\boldsymbol{\omega}} \in \mathbb{R}^3\) denoting angular velocity and angular acceleration, respectively.
Alternatively, orientation can be represented as a quaternion \(\mathbf{Q} \in \mathbb{R}^4\).
Subscripts indicate entities or coordinate components, while superscripts denote reference frames which are omitted when identical to the body frame.
We use the labels `\textit{b}', `\textit{o}', and `\textit{ee}' to refer to the robot base, object, and end-effector frames, respectively.
For instance, \(\mathbf{p}_{o}^b\) denotes the position of an obstacle relative to the robot’s base frame \(b\), with \(p_{o, x}^b\) representing its \textit{x}-component.
Manipulator joint states, including positions, velocities, and accelerations, are defined as \(\mathbf{q}\), \(\dot{\mathbf{q}}\), and \(\ddot{\mathbf{q}} \in \mathbb{R}^7\), corresponding to the seven DOFs of the robotic arm.
The obstacle's size along each axis is denoted by \(\mathbf{d}_o=[d_{o,x}, d_{o,y}, d_{o,z}]\).
Finally, a binary function \(\operatorname{is\_contact}(\cdot, \cdot)\) returns 1 if two entities are in contact, and 0 otherwise.

\section{Method}
\label{sec:method}
We aim to enhance navigation efficiency in \textit{NAMO} tasks by enabling a mobile manipulator to push obstacles with its arm while allowing its base to maintain adherence to the global path and minimize deviations. 
To achieve this, we introduce a hierarchical reinforcement learning (HRL) framework that integrates object manipulation and navigation through whole-body movements, as illustrated in Fig.~\ref{fig:framework}. 
A high-level controller selects an optimal pushing strategy based on the environment and target destination, while a low-level controller executes precise motions for effective object manipulation.
The following descriptions detail the problem formulations and hierarchical controller components.


\subsection{\textit{NAMO} Task Formulation}
We consider navigation in a known static map \( M_{static} \) (i.e., floor plans) with movable obstacles whose locations and physical properties are \textit{a priori} unknown until detected through local sensing---similar to how people rely on floor plans to determine a rough path to their destination while handling unexpected obstacles along the way.
Given a goal, the robot plans a global path \( P \) on the static map \( M_{static} \), ignoring unknown obstacles, and tracks the global path using a mobile base PD controller.
Meanwhile, upon encountering the obstacle within a predefined distance (i.e., local sensing range) that obstructs the path, the robot actively interacts with the obstacle to clear the way while progressing toward the goal. 
We assume all encountered obstacles are movable and cubic, with widths and depths between \SI{0.5}{\meter} and \SI{0.8}{\meter}, heights ranging from \SI{0.6}{\meter} to \SI{1.0}{\meter}, and randomized physical properties---including mass, center of mass (CoM), and a friction coefficient---introducing variability that necessitates adaptive interaction strategies for effective manipulation.


\subsection{Robot-Obstacle Pushing Commands}
\label{sec:command}
In our hierarchical framework, a high-level policy \( \pi_{\phi}^{h} \) defines pushing strategies (i.e., commands), which a low-level policy \( \pi_{\varphi}^{l} \) executes via whole-body control.
We represent this strategy by defining the robot’s kinematic configuration relative to the obstacle as a pushing command, which regulates obstacle motion through interaction and serves as the interface between the high-level and low-level policies.

\begin{figure}[t]
  \centering
  \includegraphics[width=0.47\textwidth]{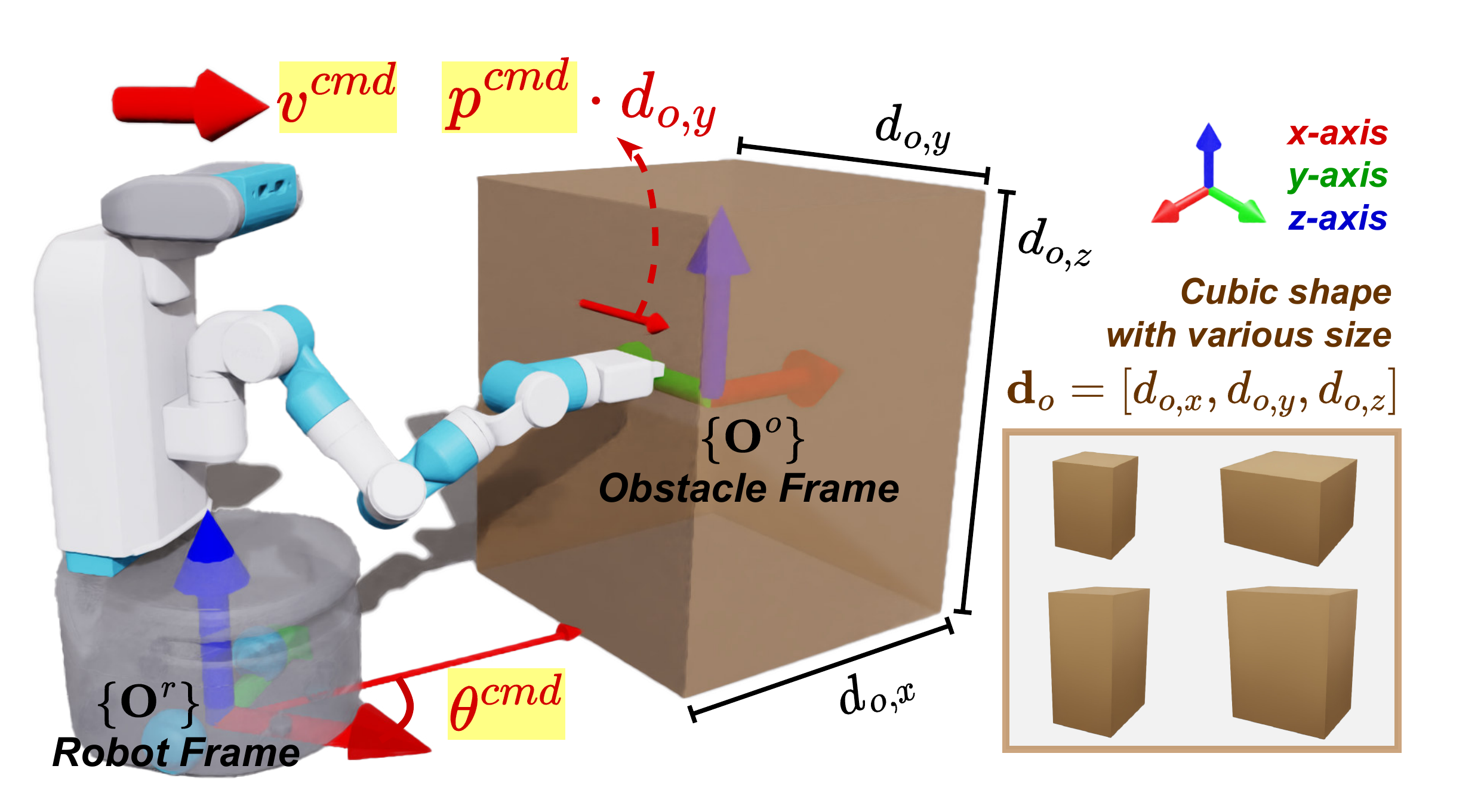}
  \caption{\small \textbf{Pushing Command}.
  The pushing command consists of \(p^{cmd}, \theta^{cmd}\), and \(v^{cmd}\), shown in red in the figure.
  \(p^{cmd}\) specifies the contact point on the obstacle contact face as \(p^{cmd} \cdot d_{o, y}\) along the obstacle's \(y\)-axis.
  \(\theta^{cmd}\) is the yaw angle of the contact face relative to the robot frame.
  \(v^{cmd}\) denotes the robot's base linear velocity.}
  \label{fig:fig_3}
  \vspace{-0.6cm}
\end{figure}

Incorporating full kinematic relationship---including all positional and motion-related variables---into the command space poses significant challenges, as the high-dimensional space increases learning complexity and complicates kinematic feasibility validation. 
To address this, we select key components that can simplify the command structure while preserving essential aspects of robot-obstacle interaction.
Specifically, the pushing command \(\mathbf{c} \in \mathbb{R}^3\) is composed of three components: $(p^{cmd}, \theta^{cmd}, v^{cmd})$, as depicted in Fig.~\ref{fig:fig_3}.
Here, $p^{cmd}$ is lateral contact position on the obstacle's contact face; $\theta^{cmd}$ denotes the yaw angle of the obstacle’s contact area relative to the robot’s frame; and $v^{cmd}$ indicates the mobile base's forward velocity during interaction.

While modeling the interaction-pushing command $\mathbf{c}$ in the low-dimensional space may cause tilting or loss of contact, the low-level controller, trained to account for these factors by generating appropriate whole-body motions based on the object's physical properties, compensates by adaptive movements, such as modifying contact height.


\subsection{A Low-Level Policy}
\label{sec:low}
The low-level policy \( \pi_{\varphi}^{l} \) enables the robot to execute whole-body movements for satisfying pushing commands $\mathbf{c}$, dynamically adapting to object tilting while maintaining stable contact through its end-effector. 
The following description details the RL formulation for the low-level policy.

\begin{table}[t]
\centering
\begin{threeparttable}
\renewcommand{\thetable}{TABLE \Roman{table}}
\captionsetup{labelformat=empty, labelsep=none}
\caption{
\small
\centering
\thetable: Reward Functions for $\mathcal{R}^{l} = \mathcal{R}^{\text{cmd}} + \mathcal{R}^{\text{stable}} + \mathcal{R}^{\text{areg}}$
}
\renewcommand{\arraystretch}{1.3}
\begin{tabular}{>{\centering\arraybackslash}p{0.13\linewidth}|>{\centering\arraybackslash}p{0.7\linewidth}}
\hline
\textbf{Reward} & \textbf{Function Expression} \\
\hline
\multicolumn{2}{c}{Command Tracking Rewards: $\mathcal{R}^{\text{cmd}} = \sum_{i=0}^{2} r^{\text{cmd}}_{i}$}
\\
\hline
$r^{\text{cmd}}_{0}$       & $k^l_0 \, exp(-2 \, |p^{cmd}-p^{o}_{ee, y}/d_{o, y} |\ /\ 0.1)$
\\
$r^{\text{cmd}}_{1}$       & $k^l_1 \, exp(-|\theta^{cmd}-\theta_{area, z}^{b}|\ /\ 0.1)$   \\
$r^{\text{cmd}}_{2}$       & $k^l_2 \, exp(-2 \, |v^{cmd}-v_{b, x} |\ /\ 0.1)$
\\
\hline
\multicolumn{2}{c}{Obstacle Stability Rewards: $\mathcal{R}^{\text{stable}} = \sum_{i=0}^{4} r^{\text{stable}}_{i}$}
\\
\hline
$r^{\text{stable}}_{0}$       & $k^l_{3} \, \operatorname{is\_contact}\left(\text{ee}, \text{obstacle}\right)$
\\
$r^{\text{stable}}_{1}$       & $k^l_{4} \, (1-\tanh(2 \, \cos^{-1}\left( \left| \textbf{Q}_{ee}^{b^{-1}} \cdot \textbf{Q}_{o}^{b} \right| \right) /\ 0.1))$
\\
$r^{\text{stable}}_{2}$       & $- k^l_{5} \, \operatorname{is\_contact}(\text{base}, \text{obstacle})$
\\
$r^{\text{stable}}_{3}$       & $- k^l_{6} \, \tanh{(|\theta^{b}_{o, y}| \ /\ 0.1)}$
\\
$r^{\text{stable}}_{4}$       & $- k^l_{7} \, (||\dot{v}^{b}_{o, x}||_2 + ||\dot{v}^{b}_{o, y}||_2)$
\\
\hline
\multicolumn{2}{c}{Action Regularization Rewards: $\mathcal{R}^{\text{areg}} = \sum_{i=0}^{2} r^{\text{areg}}_{i}$}
\\
\hline
$r^{\text{areg}}_{0}$       & $- k^l_{8} \, ||\ddot{\textbf{q}}||_2$
\\
$r^{\text{areg}}_{1}$       & $- k^l_{9} \, ||\dot{\textbf{v}}_{b}||_2$
\\
$r^{\text{areg}}_{2}$       & $- k^l_{10} \, ||\mathbf{a}_{t}-\mathbf{a}_{t-1}||_2$
\\
\hline
\end{tabular}
\label{table:ll_reward}
\begin{tablenotes}
\setlength{\itemindent}{-0.3cm}
\item[] \textbullet\; $\mathcal{R}^{l}$: a reward function for a low-level policy (refer to Sec.~\ref{sec:low})
\item[] \textbullet\; $\theta_{area, z}^{b}$: a yaw angle of an obstacle’s contact face in a robot frame
\item[] \textbullet\; $k^l_{0, 1, ...,10}$: non-negative coefficients 
\end{tablenotes}
\end{threeparttable}
\vspace{-0.8cm}
\end{table}

\subsubsection{Problem Formulation}
We formulate the obstacle-pushing problem as a Markov Decision Process (MDP), represented by the tuple \( (\mathcal{S}, \mathcal{A}, \mathcal{T}, \mathcal{R}, \rho_0, \gamma) \), where \( \mathcal{S} \) denotes the state space, \( \mathcal{A} \) the action space, \( \mathcal{T}: \mathcal{S} \times \mathcal{A} \rightarrow \mathcal{S} \) the state transition function, and \( \mathcal{R}: \mathcal{S} \times \mathcal{A} \rightarrow \mathbb{R} \) the reward function.
The term \( \rho_0 \) represents the initial state distribution, and \( \gamma \in [0, 1) \) is the discount factor.
The objective of RL is to maximize the expected cumulative reward, given by:
\begin{equation}
J(\varphi) = \mathbb{E}_{\mathbf{c}_t \sim P(\mathbf{c})} \left[ \mathbb{E}_{(\mathbf{s}, \mathbf{a}) \sim \pi^{l}_{\varphi}} \left[ \sum_{t=0}^{T} \gamma^t \mathcal{R}^{l}(\mathbf{s}_t, \mathbf{a}_t | \mathbf{c}_t) \right] \right]
\label{eq:RL}
\end{equation}
where \(\varphi\) denotes the low-level policy parameters to be optimized, and \( \mathbf{c}_t \) represents commands sampled at each time step \(t\) from an adaptively updated command distribution \( P(\mathbf{c}) \).
Please refer to Sec.~\ref{sec:curriculum} for this update procedure.
At the start of each episode, the robot joints are initialized to a predefined configuration, and the obstacle is randomly placed at a fixed distance from the robot's base.

\subsubsection{A Policy \& Reward Functions}
\label{sec:ll-policy-and-rwd}
The low-level policy \(\pi^l_{\varphi} \) receives as input the vector \( [ \mathbf{o}^{l}, \mathbf{c}, \mathbf{x} ] \), where \( \mathbf{o}^{l}\) represents the low-level observation.
This observation is structured as \( [ \mathbf{q}, v_{b,x}, \omega_{b,z}, \mathbf{p}^b_{ee}, \boldsymbol{\theta}^{b}_{ee}, \mathbf{p}^b_o, \boldsymbol{\theta}^{b}_{o}, \mathbf{d}_{o}, \mathbf{a}_{t-1} ] \).
The pushing command is given by \(\mathbf{c}\), while the privileged information \( \mathbf{x} \) consists of \([ \operatorname{is\_contact}(ee, obstacle), \, m_o, \, \mathbf{CoM}_o, \, \mu_o ] \), including the contact state between the end-effector and the obstacle, along with the obstacle's physical properties.
Property ranges and descriptions are detailed in~\ref{table:property}.
During training, ground-truth privileged information is used, whereas deployment relies on estimated values, as mentioned in Sec.~\ref{sec:priv_estimate}.
The low-level reward function \(\mathcal{R}^{l}\) consists of three main terms: command tracking rewards $\mathcal{R}^{\text{cmd}}$ which encourages adherence to the command; stability rewards $\mathcal{R}^{\text{stable}}$ which maintains contact and prevents obstacle rollover; and action regularization rewards $\mathcal{R}^{\text{areg}}$ which discourages excessive motion.
Maintaining contact is crucial, as inconsistent forces can lead to obstacle tilting or slipping, resulting in unstable interactions~\cite{agarwal1997nonholonomic, lynch1996stable, stuber2020let}.
\ref{table:ll_reward} provides detailed formulations.
We omit the time step \( t \) notation hereafter.

The action \( \mathbf{a}_t \in \mathbb{R}^8 \), comprising six dimensions for the arm and two for the base, is generated by the policy.
The arm action is defined by the desired twist of the end-effector \( [\mathbf{v}, \, \boldsymbol{\omega}]^{r}_{ee} \in \mathbb{R}^6 \), which is converted into joint velocities using a differential inverse kinematics (IK) controller~\cite{wampler1986manipulator}, resulting in joint velocities \(\dot{\mathbf{q}}\), used to update the joint positions at each time step as \( \mathbf{q} + \dot{\mathbf{q}} \cdot \Delta T\).
The base action is defined as \( [v_{b,x}, \, \omega_{b,z}] \in \mathbb{R}^2 \), reflecting the base's differential-drive kinematics, which naturally use linear and angular velocities to describe planar motion.
Both arm and base actions are then sent to joint impedance controllers for torque computation.

\subsubsection{Adaptive Command Sampling}
\label{sec:curriculum}
Learning over a large command space from scratch is challenging.
While pushing an obstacle’s center is straightforward, handling varied contact positions introduces complexity.
Sampling from an excessively large command space may yield kinematically infeasible commands, hindering learning.
To address this, we adopt a grid adaptation rule~\cite{margolis2024rapid}, which progressively expands the command sampling range for \( p^{cmd} \) and \( \theta^{cmd} \) within the pushing command space $\mathbf{c}$.
The velocity component \( v^{cmd} \) maintains a fixed sampling range, allowing the policy to focus more on interaction-related parameters.
The sampling range updates based on command tracking rewards \( r_{i}^{cmd} \).
When all rewards exceed their thresholds \( \gamma_{i} \, (i=0, 1, 2)\), the command space expands, and the probability distribution is updated as follows:
\begin{equation}
P_{N+1}(p^{cmd}, \theta^{cmd} \oplus \Delta) =
\begin{cases} 
\frac{1}{|A \cup \Delta|} \mathbb{U}_{A \cup \Delta},  \hspace{0.84cm} \text{if } \forall r_i > \gamma_{i}, \\
P_N(p^{cmd}, \theta^{cmd}),  \hspace{0.5cm} \text{otherwise},
\end{cases}
\label{eq:cmd}
\end{equation}
where \( N \) is the episode number and \( \oplus \) represents the Minkowski sum, expanding the command space by incorporating neighboring regions.
\( A \) and \( \Delta \) refer to the command space before expansion and the newly added region, respectively, with \( \mathbb{U}_{A\cup \Delta} \) representing a uniform distribution over the expanded space \( A \cup \Delta \).

\begin{table}[t]
\centering
\renewcommand{\arraystretch}{1.2}
\renewcommand{\thetable}{TABLE \Roman{table}}
\captionsetup{labelformat=empty, labelsep=none}
\caption{\small \thetable: Obstacle Physical Properties Ranges and Descriptions}
\label{tab:2} 
\begin{tabular}{>{\centering\arraybackslash}p{1.5cm}|>{\centering\arraybackslash}p{2.0cm}|>{\centering\arraybackslash}p{4.0cm}} \hline
\textbf{Term (dim.)}   & \textbf{Range (Unit)}  & \textbf{Description}\\ \hline
$m_o \, (1)$  & (5.0, 30.0) (\si{kg}) & \text{obstacle mass} \\
$\mathbf{CoM}_o \, (3)$  & (-0.4, 0.4) $(-)$ & \text{shifted obstacle CoM} \\
${\mu}_o \, (1)$  & (0.2, 0.6) $(-)$ & obstacle kinetic friction coefficient
\\ \hline
\end{tabular}
\begin{tablenotes}
\setlength{\itemindent}{-0.3cm}
\item[] \textbullet\; An obstacle CoM is set to $\mathbf{CoM}_o \odot \mathbf{d}_o$, where $\odot$ indicates element-wise multiplication and $\mathbf{d}_o$ denotes the obstacle size.
\item[] \textbullet\; During evaluation, these parameter ranges are expanded by 10\%.
\label{table:property}
\end{tablenotes}
\vspace{-0.5cm}
\end{table}

\subsection{A High-Level Policy}
\label{sec:high}
The high-level policy \(\pi^{h}_{\phi}\) generates optimal pushing commands $\mathbf{c}$, allowing the robot's base to follow the planned path with minimal deviation while actively clearing obstacles with its arm and base motions.
It also considers environmental constraints, including structures around the robot in the static map \( M_{static} \), to recognize where the obstacle can be pushed. 
The following is the RL formulation of the high-level policy.

\subsubsection{Problem Formulation}
We formulate the pushing command generation problem as an MDP for decision-making, where the RL policy aims to maximize the expected cumulative reward, similar to the low-level policy's objective:
\begin{equation}
J(\phi) = \mathbb{E}_{(\mathbf{s}_t, \mathbf{c}_t) \sim \pi^h_{\phi}} \left[ \sum_{t=0}^T \gamma^t \mathcal{R}^h(\mathbf{s}_t, \mathbf{c}_t \mid \pi^l_{\varphi}) \right].
\end{equation}
Unlike the low-level policy, which directly outputs robot actions, the high-level policy generates pushing commands \( \mathbf{c}_{t} \), executed by the low-level policy \( \pi^{l}_{\varphi} \).
During high-level policy training, the low-level policy remains frozen.

\subsubsection{Privileged Knowledge Distillation}
\label{sec:priv_estimate}
Efficient obstacle pushing requires accurate knowledge of physical properties and the current contact state, which are typically considered privileged information.
As they are generally unavailable during deployment, we train a network \(e^{\text{x}}\) online during high-level policy training to estimate both contact states and obstacle physical properties based on the low-level observation history.
The network learns to predict this privileged information using the following loss function:
\begin{equation}
\mathcal{L}_{adap} = \mathcal{L}_{contact} + \mathcal{L}_{prop},
\end{equation}
where \(\mathcal{L}_{contact}\) represents the Binary Cross-Entropy (BCE) loss for contact prediction and \(\mathcal{L}_{prop}\) denotes the Mean Squared Error (MSE) loss for physical property estimation.

Training \(e^{\text{x}}\) using low-level observation history, where high- and low-level policies rely on ground truth privileged information to generate policy outputs, poses challenges during deployment due to data distribution shifts.
In early interactions, the estimated privileged information \(\hat{\textbf{x}}\) is highly uncertain, causing deviations from trained behavior.
This discrepancy alters observations and induces distribution shifts between training and deployment.
To mitigate this issue, we adopt a progressive integration strategy that gradually transitions policies from using \(\textbf{x}\) to relying entirely on \(\hat{\textbf{x}}\).
During training, the privileged input \(\textbf{x}^{comb}\) is defined as:
\begin{equation}
\textbf{x}^{comb} = (1-\alpha) \cdot \textbf{x} + \alpha \cdot \hat{\textbf{x}},
\end{equation}
where \(\alpha\) increases linearly from 0 to 1 over the course of training.
Initially, \(\textbf{x}\) provides stable supervision for policy optimization under ideal conditions.
As training progresses, reliance on \(\hat{\textbf{x}}\) increases, allowing the policy to adapt to deployment conditions and align the training and deployment distributions.

\begin{table}[t]
\centering
\begin{threeparttable}
\renewcommand{\thetable}{TABLE \Roman{table}}
\captionsetup{labelformat=empty, labelsep=none}
\caption{
\centering
\small
\thetable: Reward Functions for $\mathcal{R}^{h} = \mathcal{R}^{\text{path}} + \mathcal{R}^{\text{safe}} + \mathcal{R}^{\text{creg}}$
}
\renewcommand{\arraystretch}{1.3}
\begin{tabular}{>{\centering\arraybackslash}p{0.13\linewidth}|>{\centering\arraybackslash}p{0.7\linewidth}}
\hline
\textbf{Reward} & \textbf{Function Expression} \\
\hline
\multicolumn{2}{c}{Path Tracking Rewards: $\mathcal{R}^{\text{path}} = \sum_{i=0}^{2} r^{\text{path}}_{i}$}
\\
\hline
$r^{\text{path}}_{0}$       & $k^h_{0} \, exp(-|\theta^{b}_{\mathbf{g}_{0}, z}|\ /\ 0.1)$ \\
$r^{\text{path}}_{1}$       & $k^h_{1} \, exp(-|(1 - v_{b, x}\ /\ v^{max}_{b, x})|\ /\ 0.1)$   \\
$r^{\text{path}}_{2}$       & $k^h_{2} \, \mathbb{I}\left(\min_i |d_i| > d_{\text{thr}} \wedge \; \operatorname{sign}(d_i) = \operatorname{sign}(d_1), \forall i \right)$ \\
\hline
\multicolumn{2}{c}{Safe Operation Rewards: $\mathcal{R}^{\text{safe}} = \sum_{i=0}^{1} r^{\text{safe}}_{i}$}
\\
\hline
$r^{\text{safe}}_{0}$       & $- k^h_{3} \, \operatorname{is\_contact}(\text{base}, M_{static})$ \\
$r^{\text{safe}}_{1}$       & $- k^h_{4} \, \operatorname{is\_contact}(\text{obstacle}, M_{static})$ \\
\hline
\multicolumn{2}{c}{Command Regularization Rewards: $\mathcal{R}^{\text{creg}} = \sum_{i=0}^{1} r^{\text{creg}}_{i}$}
\\
\hline
$r^{\text{creg}}_{0}$       & $\mathcal{R}^{stable} + \mathcal{R}^{areg}$ \\
$r^{\text{creg}}_{1}$       & $- k^h_{5} \, ||\mathbf{c}_{t} - \mathbf{c}_{t-1}||_{2}$ \\
\hline
\end{tabular}
\label{table:hl_reward}
\begin{tablenotes}
\setlength{\itemindent}{-0.3cm}
\item[] \textbullet\; $\mathcal{R}^{h}$: a reward function for a high-level policy (refer to Sec.~\ref{sec:high})
\item[] \textbullet\; $k^h_{0, 1, ...,5}$: non-negative coefficients 
\item[] \textbullet\; $\mathbb{I}$: an indicator returning 1 if the condition is met, and 0 otherwise.
\item[] \textbullet\; $d_{i} (i=0, ... , 3)$: distances from the vector connecting the robot base to the local goal \(\mathbf{g}_{1}\) to each ground-contacting corner of the obstacle; sign indicates the corner's side relative to this vector
\item[] \textbullet\; $d_{\text{thr}}$: a clearance threshold for the base to move without collision
\end{tablenotes}
\end{threeparttable}
\vspace{-0.6cm}
\end{table}
\subsubsection{A Policy \& Reward Functions}
The high-level policy takes a vector \([\mathbf{o}^{h}, \mathbf{o}^{l}, \textbf{x}^{comb}] \) as input.
The high-level observation \(\mathbf{o}^{h}\) includes \(\boldsymbol{l}_{map} \), an encoded representation of the local map \(M_{local}\), which is extracted from \(M_{static}\) and processed through the map encoder \( e^{l} \).
The local map covers a \(4 \times 4 \, \text{m}^2\) area centered \(2 \, \text{m}\) ahead of the robot base, with a resolution of \(\SI{0.05}{m}\) per pixel.
Additionally, \(\mathbf{o}^{h}\) contains upcoming local goals \(\textbf{g}_{0}\), \(\textbf{g}_{1}\), \(\text{and}\ \textbf{g}_{2}\), indicating 2D positions \((x, y)\) on the planned path \(P\) at \(0.5 \, \text{m}, 1.0 \, \text{m},\) and \(2.0 \, \text{m}\) ahead, along with the last executed command $\mathbf{c}_{t-1}$.
The high-level reward function \( \mathcal{R}^{h} \) comprises three components: path tracking rewards $\mathcal{R}^{\text{path}}$ which rewards the robot for following local goals and relocating obstacles beyond the clearance threshold \(d_{thr}\); safe operation rewards $\mathcal{R}^{\text{safe}}$ which penalizes unintended collision with static structures; and command regularization rewards $\mathcal{R}^{\text{creg}}$ which discourages abrupt changes in commands. 
Detailed reward formulations are provided in~\ref{table:hl_reward}.

\subsection{Training Details}
We employed Isaac Sim~\cite{mittal2023orbit} for training, utilizing 1,024 parallel environments.
We updated policies at 50 Hz, while we operated joint impedance control at 100 Hz.
We used the Differential IK controller with the Damped Least Squares (DLS) method and a damping factor \( \lambda = 0.02 \).
We initialized command ranges as \( (p^{cmd}, \theta^{cmd})_{\text{init}} = (\pm 0.1, \pm 0.2) \) and expanded them incrementally by \( \Delta = (\pm 0.05, \pm 0.1) \), restricting linear velocity command ranges to \((0.1, 0.4)\) \SI{}{m}/\SI{}{s}.
We set command tracking reward thresholds to \( \gamma_{0,1,2} = [0.6, 0.6, 0.5] \).
We empirically found best-performing reward weights as \(k^l_{i=0, ..., 10}=[1.0, 1.0, 0.8, 1.5, 1.5, 10^2, 0.3, 3 \times 10^{-3}, 10^{-3}, 3 \times 10^{-3}, 3 \times 10^{-3}]\) for the low-level policy and \(k^h_{i=0, ..., 5}=[1.0, 0.8, 10^2, 10^2, 10^2, 0.2]\) for the high-level policy.
We set the clearance threshold \(d_{thr}\) to \(0.35\ \text{m}\), and we set \(v^{max}_{b, x}\) to \(0.4\) \SI{}{m}/\SI{}{s}.
We optimized both policies using the Proximal Policy Optimization (PPO) algorithm \cite{schulman2017proximal}.
\ref{table:network} provides network architecture details.

    \begin{table}[t]
\renewcommand{\arraystretch}{1.3}
\renewcommand{\thetable}{TABLE \Roman{table}}
\captionsetup{labelformat=empty, labelsep=none}
\caption{\small \thetable: Network Architectures and Input-Output Specifications}
\label{tab:2} 
\begin{tabular*}{\columnwidth}{p{0.33cm}|p{2.48cm}|p{3.1cm}|p{0.5cm}} \hline
\textbf{NN.}   & \textbf{Hidden Layers}  & \textbf{Inputs (dim.)} & \textbf{Outputs} \\ \hline
$\pi_{\varphi}^l$  & [256, 128, 64] & $\mathbf{o}^{l} (32)\ \vert\ \mathbf{x} (6)\ \vert\ \mathbf{c} (3)$ & $\mathbf{a} (8)$  \\
$\pi_{\phi}^h$  & [256, 128, 64] & $\mathbf{o}^{h} (25)\ \vert\ \mathbf{o}^{l} (32)\ \vert\ \mathbf{x}^{comb} (6)$ & $\mathbf{c} (3)$  \\
$e^{\text{x}}$  & LSTM + [128, 64, 32] & $\mathbf{o}^l (32)$ & $\hat{\mathbf{x}} (6)$  \\
$e^{l}$  & CNN + [128, 64. 32] & $M_{local} (80 \times 80)$ & $\boldsymbol{l}_{map} (16)$  \\
\hline
\end{tabular*}
\label{table:network}
\vspace{-0.5cm}
\end{table}

\begin{table}[H]
\renewcommand{\arraystretch}{1.15}
\renewcommand{\thetable}{TABLE \Roman{table}}
\captionsetup{labelformat=empty, labelsep=none}
\caption{\small \thetable: Low-Level Command Tracking Errors and Contact Rate}
\label{table:ll_policy} 
\begin{tabularx}{1.0\columnwidth}{>{\centering\arraybackslash}m{0.7cm}|>{\centering\arraybackslash}m{1.34cm}|>{\centering\arraybackslash}m{1.34cm}|>{\centering\arraybackslash}m{1.34cm}|>{\centering\arraybackslash}m{1.7cm}}
\hline
\textbf{} & \(p^{cmd}\) error (-) \((\downarrow)\) & \(\theta^{cmd}\) error (rad) \((\downarrow)\) & \(v^{cmd}\) error (\SI{}{m}/\SI{}{s}) \((\downarrow)\) & Contact Rate (\%) \((\uparrow)\)  \\ \hline
\text{LL-P} & 0.129 $\pm$ 0.041 & 0.169 $\pm$ 0.065 & 0.079 $\pm$ 0.027 & 89.219 $\pm$ 8.935 \\ \hline
\text{LL+P} & \textbf{0.055} $\pm$ \textbf{0.029} & \textbf{0.122} $\pm$ \textbf{0.059} & \textbf{0.049} $\pm$ \textbf{0.027} & \textbf{96.708} $\pm$ \textbf{2.621} \\ \hline
\end{tabularx}
\begin{tablenotes}
\setlength{\itemindent}{-0.3cm}
\item[] \textbullet\; P: Privileged information (refer to Sec.~\ref{sec:exp-ll})
\end{tablenotes}
\vspace{-0.4cm}
\end{table}
\begin{table*}[t]
\centering
\renewcommand{\arraystretch}{1.15}
\renewcommand{\thetable}{TABLE \Roman{table}}
\captionsetup{labelformat=empty, labelsep=none}
\caption{\small \thetable: Navigation Among Movable Obstacles (\textit{NAMO}) Experimental Results on Map 1 and Map 2}
\label{tab:map1_result} 
\begin{threeparttable}
\begin{tabularx}{0.8\textwidth}{>{\centering\arraybackslash}p{1.2cm}|>{\centering\arraybackslash}p{3.8cm}|>{\centering\arraybackslash}p{3.8cm}|>{\centering\arraybackslash}p{3.8cm}}
\hline 
\hline
\textbf{Methods} & \textbf{SR} (\%) \((\uparrow)\) & \textbf{SPL} (-) \((\uparrow)\) & \textbf{SCT} (-) \((\uparrow)\) \\ \hline
\hline
\multicolumn{4}{c}{Map 1 (evaluation with random obstacle numbers of 2, 4, and 6, presented in that order)} \\ \hline
\text{CA} & 92.00 / 72.00 / 52.00 & 84.43 / 63.86 / 44.23 & 77.94 / 56.59 / 37.82 \\ \hline
\text{AA~\cite{ellis2022navigation}} & 96.00 / 89.33 / 78.67 & 88.61 / 80.81 / 69.35 & 82.38 / 71.99 / 60.59 \\ \hline
\text{Ours-P} & 86.67 / 74.00 / 64.67 & 85.41 / 71.09 / 62.09 & 85.25 / 67.35 / 55.80 \\ \hline
\text{Ours-H} & 94.00 / 84.67 / 76.00 & 90.81 / 81.63 / 73.46 & 91.16 / 78.10 / 73.46 \\ \hline
\text{Ours} & \textbf{96.67} / 91.33 / 85.33 & 93.39 / 88.38 / 82.62 & \textbf{95.89} / 88.26 / 80.28 \\ \hline 
\text{Ours+G} & \textbf{96.67} / \textbf{92.00} / \textbf{87.33} & \textbf{93.53} / \textbf{89.12} / \textbf{84.64} & 95.86 / \textbf{89.12} / \textbf{81.96} \\ \hline
\multicolumn{4}{c}{Map 2} \\ \hline
\text{CA} & 94.00 / 88.00 / 80.00 & 85.97 / 78.03 / 68.36 & 75.75 / 66.99 / 57.18 \\ \hline
\text{AA~\cite{ellis2022navigation}} & 95.33 / 89.33 / 84.67 & 87.56 / 81.36 / 75.43 & 77.44 / 70.32 / 62.53 \\ \hline
\text{Ours-P} & 88.67 / 84.67 / 76.67 & 87.53 / 83.37 / 75.56 & 84.71 / 78.12 / 67.93 \\ \hline
\text{Ours-H} & 82.00 / 70.67 / 57.33 & 80.81 / 69.55 / 56.43 & 78.35 / 65.18 / 50.97 \\ \hline
\text{Ours} & \textbf{98.67} / 96.67 / 92.67 & \textbf{97.70} / 95.79 / 91.42 & \textbf{96.80} / 92.74 / 86.79 \\ \hline
\text{Ours+G} & \textbf{98.67} / \textbf{97.33} / \textbf{94.67} & 97.69 / \textbf{96.43} / \textbf{93.83} & 96.76 / \textbf{93.36} / \textbf{89.18} \\ \hline
\hline
\end{tabularx}
\begin{tablenotes}
\setlength{\itemindent}{-0.3cm}
\item[] \textbullet\; \textbf{SR}: success rate, \textbf{SPL}: success-weighted path length (see Eq.~\ref{eq:spl}), \textbf{SCT}: success-weighted completion time (see Eq.~\ref{eq:sct}).
\end{tablenotes}
\end{threeparttable}
\vspace{-0.35cm}
\end{table*}


\section{Experiments}
\label{sec:exp}
We conducted two simulation experiments to evaluate our approach: (1) analyzing the impact of privileged information on low-level policy performance and (2) assessing navigation performance in \textit{NAMO} tasks across varying obstacle densities measured by success rate, path length, and completion time.

\subsection{Effect of Privileged Information on Low-Level Policy}
\label{sec:exp-ll}
To assess the impact of privileged information on low-level policy performance, we compared policies trained with (LL+P) and without (LL-P) privileged information. 
\ref{table:ll_policy} presents tracking errors for each pushing-command term and the consistency of maintaining contact with the obstacle.

The LL+P policy exhibits improved command adherence, with reduced tracking errors across all command dimensions. 
Specifically, lower \(p^{cmd}, \theta^{cmd}\), and \(v^{cmd}\) errors demonstrate enhanced precision in executing high-level commands, ensuring more reliable robot-obstacle interaction.
In addition, it maintains more stable and prolonged contact with the obstacle, reinforcing consistent interaction. 
These results validate that privileged information improves both control accuracy and interaction stability.

\begin{figure}[t]
  \centering
  \includegraphics[width=0.48\textwidth]{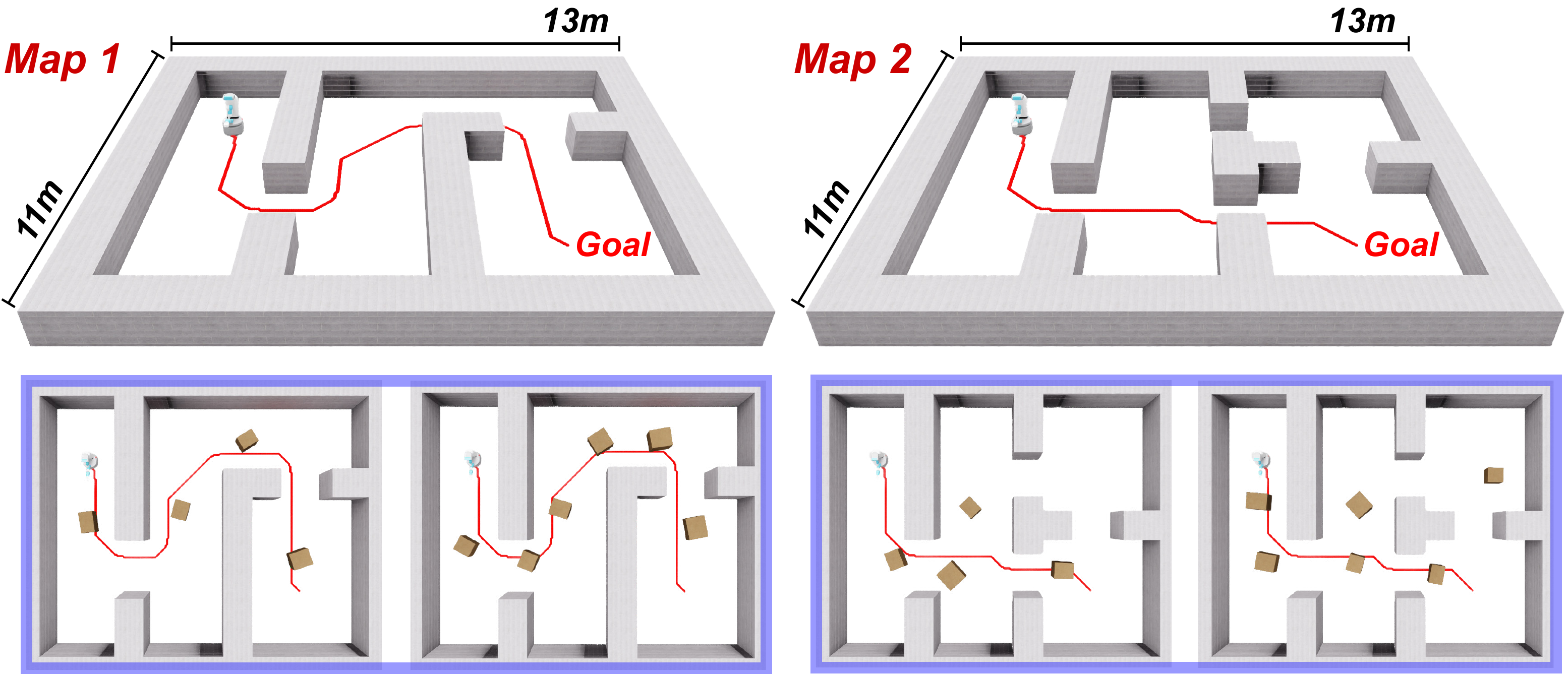}
  \caption{\small \textbf{Evaluation Maps}.
  These maps are distinct from the training environments, with fixed start and goal positions.
  The red line represents the planned path \(P\), computed using the A* algorithm.
  The bottom row illustrates example configurations for each map with 4 and 6 obstacles.}
  \label{fig:eval_map}
  \vspace{-0.5cm}
\end{figure}

\begin{figure}[t]
  \centering
  \includegraphics[width=0.49\textwidth]{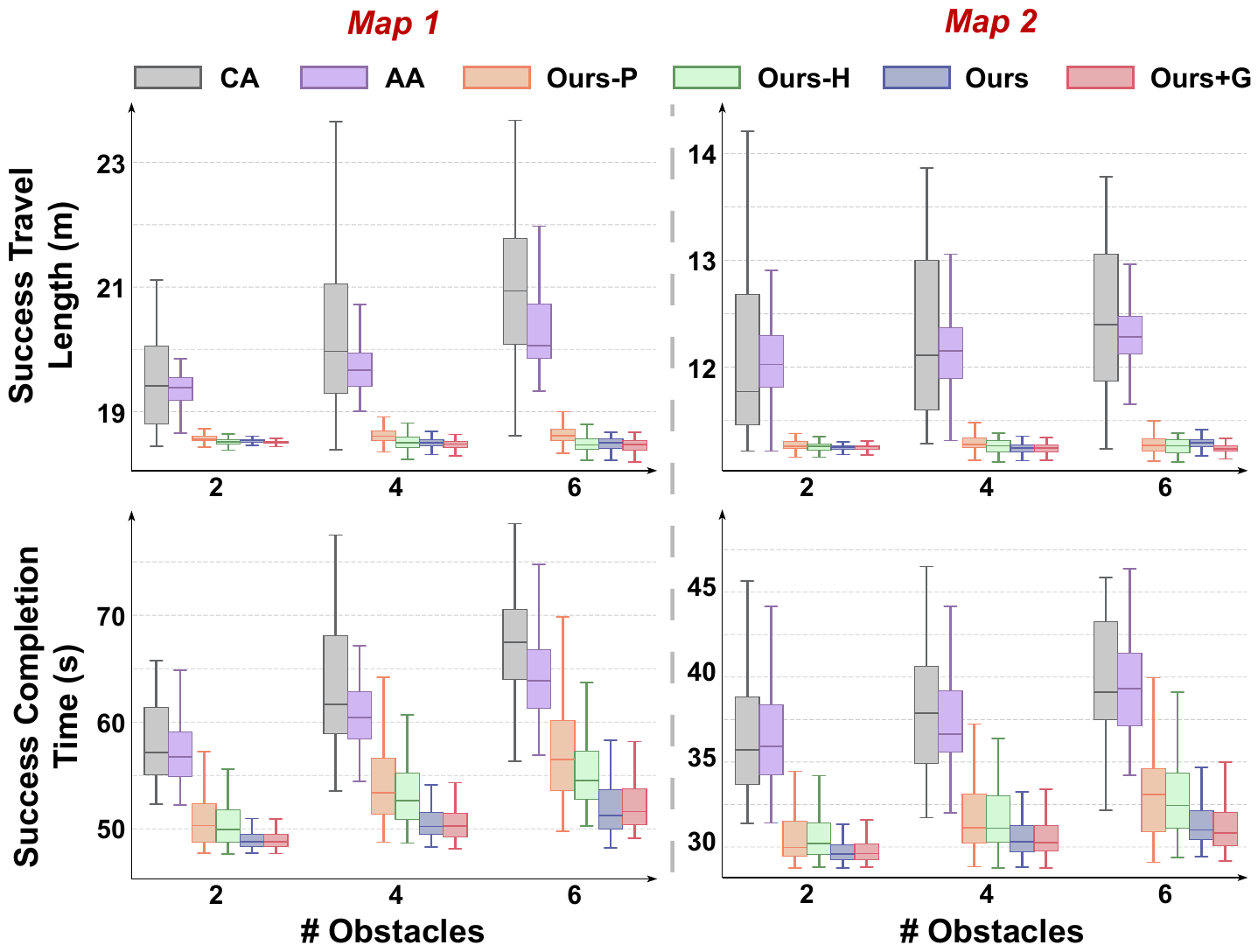}
  \caption{\small \textbf{Navigation Efficiency for Success Cases}.
  Travel length and completion time are evaluated for successful trials across methods, measured under varying obstacle counts.
  \textbf{Ours} achieves consistently shorter distances and times with lower variance, demonstrating stable and efficient performance across all cases.}
  \label{fig:boxplot}
  \vspace{-0.5cm}
\end{figure}

\subsection{Navigation Efficiency in Cluttered Environments}
\subsubsection{Experiment Settings}
We evaluated the navigation efficiency of our method on two maps, each with 2, 4, and 6 obstacles strategically placed to interfere with the planned global path, as shown in Fig.~\ref{fig:eval_map}.
For each map and obstacle count setting, we conducted 150 trials.
In each experiment, obstacle properties are randomly assigned within the ranges specified in \ref{tab:2}.
The start and goal positions are fixed, and the global path $P$ is computed using the A* algorithm~\cite{hart1968formal}, considering only static walls while ignoring movable obstacles.
During navigation, if the robot encounters an obstacle within 1.5\SI{}{m} that obstructs the path, our method uses the proposed policies to push it aside.
Once the obstacle is cleared---its clearance exceeds the threshold \(d_{thr}\)---the robot reverts to tracking its planned path using base movements.
This process continues until the robot either reaches the goal or fails according to predefined criteria described below.

\subsubsection{Baselines and Metrics}
We established the following baselines along with our method's variants:
\begin{itemize}
    \item \textbf{Collision Avoidance (CA)}: Re-plans paths iteratively to avoid obstacles without physical interaction.
    \item \textbf{Axis-Aligned Pushing (AA)}~\cite{ellis2022navigation}: Re-plans paths with a straight-line pushing strategy, aligning the robot's base with the obstacle’s axis to clear the obstructed path.
    \item \textbf{Ours w/o Privileged Information (Ours-P)}: Trains hierarchical policies without privileged information.
    \item \textbf{Ours w/o Hierarchical Structure (Ours-H)}: Trains a single policy end-to-end that directly generates low-level actions \(\mathbf{a}_t\) (refer to Sec.~\ref{sec:ll-policy-and-rwd}), omitting hierarchical structure interconnected via pushing commands.
    \item \textbf{Ours}: Uses the hierarchical structure along with pushing commands and estimated privileged information in both high- and low-level policies (refer to Sec.~\ref{sec:method}).
    \item \textbf{Ours w/ GT Privileged Information (Ours+G)}: Uses ground-truth privileged information instead of estimated values through the network \(e^{\text{x}}\) (refer to Sec.~\ref{sec:priv_estimate}).
\end{itemize}

\begin{figure}[t]
  \centering
  \hspace{-0.4cm}
  \includegraphics[width=0.48\textwidth]{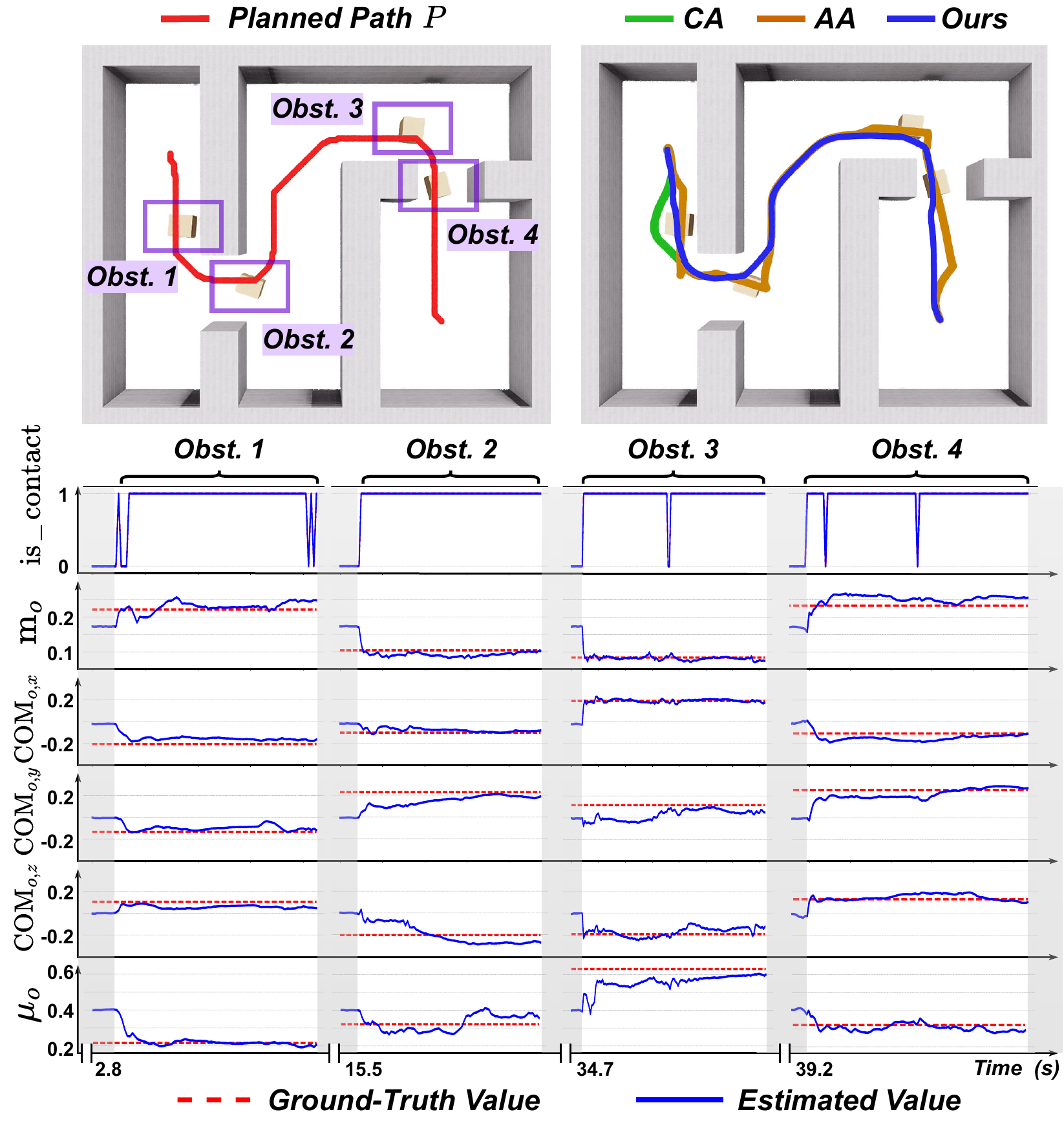}
  \caption{\small \textbf{Qualitative Results on Map 1}.
  The top-row figures shows the planned path \(P\) and the robot's traces for \textbf{CA}, \textbf{AA}, and \textbf{Ours}, where \textbf{CA} fails to find a feasible path around the second obstacle, resulting in task failure.
  The below figures present the privileged information estimated by \textbf{Ours} during obstacle interactions alongside ground-truth values. 
  Shaded gray regions indicate periods when the robot is not in contact with the obstacle.}
  \label{fig:qualitative}
  \vspace{-0.5cm}
\end{figure}

To quantify navigation efficiency, we used three metrics:
\textbf{Success Rate (SR)} is the percentage of trials in which the robot successfully reaches the goal. A trial is considered a failure if the robot exceeds a predefined time limit or if an obstacle collides with a wall or tips over. In the case of \textbf{CA}, failure additionally occurs when it cannot find a feasible path.
\textbf{Success-Weighted Path Length (SPL)}~\cite{anderson2018evaluation} measures path efficiency by balancing optimality and success, defined as:
\begin{equation}
\text{SPL} = \frac{1}{N} \sum_{i=1}^{N} S_i \frac{L_i^*}{\max(L_i, L_i^*)},
\label{eq:spl}
\end{equation}
where $S_i$ denotes success, $L_i^*$ the optimal path length, and $L_i$ the actual path length.
Similarly, \textbf{Success-Weighted Completion Time (SCT)}~\cite{yokoyama2021success} evaluates time efficiency:
\begin{equation}
\text{SCT} = \frac{1}{N} \sum_{i=1}^{N} S_i \frac{T_i^*}{\max(T_i, T_i^*)},
\label{eq:sct}
\end{equation}
where $T_i^*$ is the optimal completion time, and $T_i$ is the actual completion time.
The optimal values \(L_i^*\) and \(T_i^*\) are determined from navigation in an obstacle-free environment.

\subsubsection{Analysis of Results}
\ref{tab:map1_result} presents the navigation performance across baseline methods. 
Our approach (\textbf{Ours}), integrating interaction-based obstacle property estimation and the hierarchical structure, consistently outperforms the baselines, particularly in environments with a high number of obstacles.
\textbf{CA} frequently fails to find feasible paths in settings with many obstacles, resulting in a low success rate.
Its avoidance-based strategy results in longer trajectories, reducing navigation efficiency.
\textbf{AA} improves upon \textbf{CA} but lacks adaptability, as it ignores obstacle physical properties, leading to failed or unintended interactions.
Additionally, its reliance on base-only pushing introduces redundant movements, further lowering efficiency (see Fig.~\ref{fig:qualitative}).
\textbf{Ours-P}, which does not estimate obstacle properties, exhibits significantly lower success rates and reduced efficiency compared to \textbf{Ours}, underscoring the importance of incorporating obstacle properties for adaptive manipulation.
\textbf{Ours-H}, which removes the hierarchical structure, struggles with inefficient and ineffective learning due to the large search space, converging into suboptimal solutions and leading to poorer navigation performance, consistent with findings in~\cite{tan2021hierarchical}.
\textbf{Ours+G}, which uses ground-truth privileged information, predictably achieves the highest performance, serving as the upper bound for our approach. 
Through the small performance gap between \textbf{Ours} and \textbf{Ours+G}, it can be inferred that \textbf{Ours} accurately estimates the necessary privileged information for efficient obstacle manipulation.
Fig.~\ref{fig:qualitative} shows a qualitative result confirming the accuracy of \textbf{Ours}, as it reliably estimates obstacle properties in real time while tracking the planned global path with minimal deviation.
This experiment confirms that integrating property estimation with a hierarchical policy framework enhances robustness and efficiency in \textit{NAMO} tasks.

Fig.~\ref{fig:boxplot} presents an analysis of navigation efficiency based on travel length and completion time for successful trials in each experimental case.
\textbf{CA} and \textbf{AA} result in longer travel distances with high variance, indicating frequent detours and inefficient obstacle interactions, respectively. 
In contrast, \textbf{Ours} consistently achieves shorter travel lengths and completion times across all obstacle densities, with minimal variance, demonstrating robust and efficient navigation.
These results confirm that \textbf{Ours} not only enhances success rates but also improves efficiency across diverse cluttered environments, validating the effectiveness of our approach.

For a more intuitive understanding of the experimental results, please refer to the accompanying supplementary video which demonstrates the obstacle manipulation process.

    \section{Conclusion}
We proposed a hierarchical reinforcement learning (HRL) framework for online Navigation Among Movable Obstacles (\textit{NAMO}) using a mobile manipulator with a 7-DOFs robotic arm. 
Our approach integrates interaction-based obstacle property estimation with structured pushing strategies, where the high-level policy generates pushing commands, and the low-level policy facilitates accurate and stable execution.
Our extensive evaluations confirm that the proposed approach significantly enhances navigation efficiency compared to baseline methods.
Although our method has shown promising results, it currently assumes cubic-shaped, movable obstacles. 
To address such an assumption, future work will extend object geometries to diverse shapes, such as cylinders and general objects like chairs, and explore environments with both movable and immovable obstacles. 
We also aim to evaluate real-world robustness, considering sensor noise and sim-to-real gaps.

    {
        \small
        \bibliographystyle{ieee}
        \bibliography{./ref}
    }
\end{document}